\documentclass{article}

\usepackage[greek,english]{babel}
\usepackage[utf8x]{inputenc}
\usepackage{amsmath}
\usepackage{graphicx}
\usepackage{float}
\usepackage{attrib}
\usepackage{url}
\def\agreement{\mathrm{agreement}}
\def\NCD{\mathrm{NCD}}
\def\C{\mathrm{C}}

\title{Unsupervised authorship attribution}
\author{David Fifield, Torbjørn Follan, Emil Lunde}

\begin{document}

\maketitle

\renewcommand{\abstractname}{}
\begin{abstract}
We describe a technique for attributing parts of a written text to a set of unknown authors.
Nothing is assumed to be known a priori about the writing styles of potential authors.
We use multiple independent clusterings of an input text to identify parts that are similar and dissimilar to one another.
We describe algorithms necessary to combine the multiple clusterings into a meaningful output.
We show results of the application of the technique on texts having multiple writing styles.
\end{abstract}

\section{Problem statement}

Accept as input a written text and a number $n$ of presumed authors.
For each small segment of the text---for example, for each word or line---output
a weight assignment for the labels $1,\ldots,n$, with the weights summing to $1$.

Nothing is previously assumed about the authors---there are no training samples of writing by potential authors.
The author labels in the output are just integers, not tied to any real personal identity.
The weights on each segment may be interpreted as a probability distribution over authors;
for example a segment may appear as 15\% Unknown Author~1, 80\% Unknown Author~2, and 5\% Unknown Author~3.
The output labels are arbitrary: any permutation of labels that is applied consistently
across all segments is as good as any other.

The input text is an ordinary written document in a human language such as English.
The text doesn't have explicit section markings or other divisions; from the point of view of the algorithm it is just a long sequence of words.
The $n$ authors may have worked together on the document at the same time,
or the text may have been assembled using parts from different authors separated in time.

\subsection{Prior work}

There is a great deal of literature on the topic of supervised authorship attribution---extracting lexical
and other features from writing samples known to belong to different authors,
and using those to help decide the authorship of an unknown text. Stamatatos~\cite{stamatatos}
gives a survey. There is comparatively little on assigning authorship to authors
without prior knowledge of their writing styles.
Koppel et~al.~\cite{koppel} use synonym choice, among other features,
to learn authorship divisions in input texts artificially constructed from different parts of the Bible.

\section{Solution technique}

A technique for assigning $n$ labels to parts of a document suggests itself:
Break the input into fragments, cluster the fragments into $n$ clusters, and assign a label to each fragment equal to the cluster label.
The clustering may be based on, for example, stylometric features extracted from the fragments.
We expect that fragments within a cluster are similar to one another but dissimilar to fragments in other clusters.
If the similarities and dissimilarities come principally from different authors' stylistic choices,
the cluster labels will reflect authorship.

There is a problem with the technique just described:
A fragment of text needs to have a certain minimum length in order to represent an author's style.
The minimum has been found to vary with different classification methods~\cite{stamatatos},
but it should be at least several hundred words for English text.
Longer fragments convey more of an author's style, at the cost of low-resolution output:
If all fragments are 1000 words long, the output may show a change in authorship only every 1000 words.
Smaller fragments give better resolution, at the cost of less distinctive style within fragments and less robust clustering.

Our proposed solution to this problem is to repeat the naive clustering $m$ times,
each time using fragments that are shifted relative to the previous time.
Reassign cluster labels within clusterings to make the clusterings agree as much as possible.
Output the average of the $m$ relabeled clusterings.
This is a summary of the procedure:

\begin{enumerate}
\item Make $m$ copies of the input, and break each copy into fragments,
with the fragments in one copy being shifted relative to the other copies.
\item Cluster each set of fragments independently according to some similarity measure.
\item Permute the labels within each clustering to maximize pairwise clustering agreement.
\item Output the average of the relabeled clusterings.
\end{enumerate}

For example, let the fragment size be 1000 words and the shift offset be 50 words.
These parameters lead to $m=1000/50=20$ sets of fragments. The first comprises words
1--1000,1001--2000,\ldots; the second words
51--1050,1051--2050,\ldots; and so on.
We omit short fragments, for example the one composed of words 1--51 that would otherwise be in the second fragment set.
The output will be short segments of only 50 words, each with its own weight distribution.

Fragments do not necessarily have to be the same size. They should be long enough
to represent a single author's style,
but not so long that fragments tend to have more than one or two authors.
Fragments that straddle a true change in authorship tend to be assigned to one or the other author
according to how much each author contributed to the fragment.
The act of averaging fragments that start at different offsets therefore
tends to cause label weights to transition smoothly between one label and another,
at least where the change in authorship is clear.

The overall effect is that we get output fine-grained enough to localize authorship changes,
while still doing clustering over longer, stylistically meaningful fragments.
Repeated clustering introduces an additional complication, taken care of in step~3.
Because the $m$ clusterings are independent, they will not in general use the same label for the same unknown author.
Step~3 permutes labels within each clustering, as we are free to do,
to make the labels agree as much as possible.

The following sections consider different techniques for doing step~2 (clustering),
and step~3 (matching of clustering labels).

\section{Clustering techniques}

In this section we consider ways to cluster text fragments according to their stylistic similarity.
Clustering over feature vectors (Section~\ref{sec:jstylo}) is amenable to a variety of machine learning techniques
such as dimensionality reduction (Section~\ref{sec:dimreduc}); it requires some amount of natural language processing
in order to extract the features.
Compression clustering (Section~\ref{sec:compcluster}) does not deal with meaningful feature vectors,
but may be used even when it is inconvenient to extract features,
for example when the input is in a language for which natural language processing libraries are lacking.

\subsection{Stylometric feature vectors}
\label{sec:jstylo}

There are many ways to transform text into a numeric feature vector.
Typical features are things like average word length, and frequency of use of punctuation---but there exist much more sophisticated features.
Once a set of feature vectors has been produced, they may be clustered using any of a variety of general-purpose clustering algorithms.

JStylo~\cite{jstylo} is a framework for authorship attribution, usually used to evaluate a test set of documents against a training set.
We used JStylo to generate feature vectors over fragment sets that were previously generated.
We used the built-in ``WritePrints (Limited)'' feature set.
The WritePrints feature set is language-aware and specific to English,
having features not only for things like word length and character frequency,
but also parts of speech and frequency of function words~\cite{writeprints}.
We then used the machine learning framework Weka~\cite{weka} to generate $m$ sets of $n$ clusters using the EM clusterer.
Clustering of JStylo-produced feature vectors was used to get the results in Section~\ref{sec:ulysses}.

\subsection{Dimensionality reduction}
\label{sec:dimreduc}

Dimensionality reduction in the sense of clustering, is the act of reducing the potentially large number of features in a feature vector. This is used to speed up the clustering software. The algorithm we use is a random-projection-type $k$-means algorithm proven to distort the clusters by at most a factor of $2+\epsilon$ for an error parameter $\epsilon$. The algorithm we use is due to Boutsidis, Drineas, and Zouzias~\cite[Algorithm~1]{randproject}.
It uses a random scaled $\pm1$ $d\times t$ matrix to reduce $d$ dimensions to $t$ dimensions, then runs an approximate $k$-means algorithm on the projected vectors.

The parameter $t$, which is the new number of features, is independent of $d$, which is the original number of features. In our case, $d$, the number of features, was around 450--500, and $k$, the number of clusters, was $2$. We chose $\epsilon=0.2$ and $c=2$, giving $t=100$ output features.

\subsection{Compression clustering}
\label{sec:compcluster}

Clustering based on traditional stylometric features requires knowledge about the data in order to extract useful feature that can be used to compare the objects. This is a problem since there are a lot of different languages, with different properties and alphabets, that we may want to cluster.

Clustering by compression is a clustering method introduced by Cilibrasi and Vitanyi~\cite{ncd}
that doesn't require any subject-specific features or knowledge about the data. The idea is that two objects are deemed similar if one can use information about one object to compress the other. More formally the normalized compression distance between objects $x$ and $y$ is
\[
\NCD(x, y) = \frac{\C(xy) - \min(\C(x), \C(y))}{\max(\C(x), \C(y)))},
\]
where $\C(x)$ is the compressed size of object $x$, and $xy$ is the concatenation of $x$ and $y$.

The normalized compression distance allows us to cluster texts written in languages we don't have tools to extract features from.
To test this method we used the CompLearn utilities~\cite{complearn} to generate a distance matrix between fragments and the $k$-medioids algorithm to cluster them.
An example of the application of this approach appears in Section~\ref{sec:apollo}.

\section{Clustering matching algorithms}

The output of step~2 of the algorithm is a set of $n$ clusterings,
each one being a labeling of fragments.
The labels within each clustering may be permuted arbitrarily.
The job of step~3 is to find a permutation of labels within each clustering
so that the labels across clusterings agree as much as possible.

We examined two special cases of this optimization:
the case where there are two authors and multiple clusterings,
and the case where there are multiple authors and two clusterings.

\subsection{Agreement between labeled fragments}

We define the normalized \emph{agreement} between a pair of fragment sets $A$ and $B$ on a text $T$:
\[
\agreement(A, B) = \frac{1}{|T|} \sum_{f_a\in A} \sum_{f_b\in B} |f_a\cap f_b| \sum_{i=1}^n (f_a)_i (f_b)_i.
\]
$|T|$ is the length of the text, for example a count of its words.
$A$ and $B$ are fragment sets; $f_a$ and $f_b$ are individual fragments within them.
$|f_a\cap f_b|$ is the size of the intersection of $f_a$ and $f_b$; for nonoverlapping fragments it is $0$.
$(f_A)_i$ and $(f_B)_i$ are the weight assignments for author $i$ in $f_A$ and $f_B$ respectively.
Simply stated, the agreement between two fragments is the size of their intersection times the dot product of their weight vectors.
The total agreement between two sets is the sum of agreements between all fragments in $A$ and all fragments in $B$;
it is a number between $0$ and $1$.

The sum-of-pairs agreement of a sequence $M$ containing $m$ sets of fragments is the normalized sum of their pairwise agreements:
\[
\agreement(M) = \frac{1}{{m \choose 2}} \sum_{1\le i<m}\sum_{i<j\le m} \agreement(M_i, M_j).
\]
The use of a sum-of-pairs objective function has analogs in algorithms used in biology for multiple sequence alignment---there,
the algorithm must choose where to insert gaps into various sequences in order to maximize a score;
here, the algorithm's choice is over permutations of labels.

We use the agreement objective for two purposes. First, sum-of-pairs agreement is used to optimize labels of the $m$ independent fragment clusterings.
Second, single-pair agreement is used to assess the effectiveness of the algorithm's output in comparison
with some known ground truth. Examples of the latter use are given in Section~\ref{sec:results}.
In the specific case where there are two authors, and label weights are constrained to be $0$ or $1$,
optimization of single-pair agreement has a natural interpretation as a binary classification problem.
Without loss of generality, fix the labels of one fragment set in the pair.
The fragments with label $1$ are the positive instances, and those with label $2$ are the negative instances.
Then the agreement is just the normalized sum of true positives and true negatives over the total:
$\frac{1}{|T|} (\mathrm{TP} + \mathrm{TN})/(\mathrm{TP} + \mathrm{TN} + \mathrm{FP} + \mathrm{FN})$.

\subsection{Two clusterings, $n>2$ authors: maximum-weight matching}
\label{sec:two-n}

When there are only two clusterings, the problem of how to permute the two label sets for maximum agreement is an instance of maximum-weight matching.
Build a complete bipartite graph with $n$ vertices on each side, corresponding to the $n$ labels in each clustering.
For each pair $1\le i,j \le n$, assign a weight equal to
the size of the intersection between fragments labeled $i$ on the left and fragments labeled $j$ on the right;
this is the size of the contribution toward the total agreement that would be achieved if label $i$ on the left were changed to $j$.
There are algorithms to solve maximum-weight matching optimally in polynomial time.

\subsection{$m>2$ clusterings, two authors: MAX-CUT using a semidefinite programming relaxation}
\label{sec:m-two}

There are only two permutations of two labels: $(1, 2)$ and $(2, 1)$.
The choice of a permutation of labels in a fragment set therefore boils down to the choice of whether to flip the labels, or not.
Consider a two-element weight vector within a fragment $(w_1,w_2)$. Because $w_1+w_2=1$,
it is the case that flipping the labels has the same effect as flipping the weights: the weight vector becomes
$((1-w_1),(1-w_2)) = (w_2,w_1)$.
The same is true of the agreement, which is the sum of dot products of weight vectors:
\[
\agreement(1-A,B)=\agreement(A,1-B)=1-\agreement(A,B).
\]
We seek to decide, for each fragment set, whether to flip its labels or not,
so that the sum or pairwise agreements is as large as possible.

Optimizing the agreement between multiple two-author clusterings can be posed as an instance of the MAX-CUT problem.
An instance of MAX-CUT is an undirected graph $G=(V,E)$ with nonnegative edge weights.
A solution to the problem is a mapping from vertices to labels in $\{-1,+1\}$
such that the sum of weights on edges whose endpoints have different labels is maximized.
Vertices with label $-1$ correspond to one side of a graph cut; those with label $+1$ correspond to the other side,
and the objective is to maximize the weight of the edges crossing the cut.
Let an instance of MAX-CUT be given with vertex set $V=\{v_1,\ldots,v_m\}$ and edge weights $w_{ij}=w_{ji}$.
Its solution is the solution to the integer program:
\begin{align*}
\max\quad&\sum_{0\le i<n}\sum_{i<j\le m} w_{ij}\frac{1-y_iy_j}{2}\\
\mathrm{subject~to\quad}&y_i\in\{-1,+1\}\qquad\forall i\in V
\end{align*}
The objective function has the same solution---though not the same value---after
removing constant positive factors and additive terms:
$\max\sum_{0\le i<n}\sum_{i<j\le m} -w_{ij}y_iy_j$
(using payoff multipliers in $\{-1,+1\}$ rather than $\{0,1\}$).

The reduction from two-author clustering matching to MAX-CUT is achieved simply by setting
$w_{ij} = -\agreement(M_i,M_j)$. Intuitively, if the agreement is large (greater than $0.5$),
then we want $-w_{ij}y_iy_j$ to be small, meaning that $y_i$ and $y_j$ should have the same sign
(flip neither of $v_i$ and $v_j$, or flip both). If the agreement is small (less than $0.5$),
$y_i$ and $y_j$ should have opposite signs (flip one label set of the other, but not both).
More formally, we may work out a payoff function that pays $+\agreement(M_i,M_j)$ when
$y_i$ and $y_j$ have the same sign, and $-\agreement(M_i,M_j)$ when
$y_i$ and $y_j$ have opposite signs:
\begin{align*}
&\agreement(M_i,M_j)\left(\frac{1+y_iy_j}{2}\right) - \agreement(M_i,M_j)\left(1-\frac{1+y_iy_j}{2}\right)\\
=~&\agreement(M_i,M_j)\left(\frac{1+y_iy_j}{2}\right) - \agreement(M_i,M_j)\left(\frac{1-y_iy_j}{2}\right)\\
=~&\agreement(M_i,M_j)\left(\frac{1+y_iy_j-1+y_iy_j}{2}\right)\\
=~&\agreement(M_i,M_j)y_iy_j
\end{align*}
which has the same maximum as the MAX-CUT program when assigning $w_{ij} = \agreement(M_i,M_j)$.

Goemans and Williamson~\cite{maxcut} show a randomized polynomial-time $0.878$-approximate algorithm to solve
MAX-CUT by way of a semidefinite programming relaxation.
We apply this algorithm after reducing two-author clustering matching to MAX-CUT.
The output labels $y_1,\ldots,y_m \in \{-1,+1\}$ map to different sides of the cut.
The clusterings on one side of the cut have their labels flipped;
and those on the other side are left unchanged.

There are other natural interpretations of the edge weights we assign:
If we regard the labels as being in $\{0,1\}$, with aligned fragments of identical length,
then the weight is the negative of the Hamming distance between the bit strings representing the labels.
If we regard the labels as being in $\{-1,+1\}$, under the same conditions,
then the weight is the negative of the dot product or cosine similarity between the label vectors.

\subsection{$m>2$ clusterings, $n>2$ authors}
\label{sec:m-n}

We did not try to solve any matchings where there are more than two authors and more than two clusterings.
It is likely that approximate or heuristic techniques will be necessary.
We believe that the application of algorithms used for sequence alignment in biology is promising.

\section{Results}
\label{sec:results}

\subsection{\textsl{Ulysses}}
\label{sec:ulysses}

\begin{quote}
\noindent
Ineluctable modality of the visible: at least that if no more, thought
through my eyes. Signatures of all things I am here to read, seaspawn and
seawrack, the nearing tide, that rusty boot.
\attrib{Beginning of episode~3}
\end{quote}

\noindent
The novel \textsl{Ulysses} by James Joyce~\cite{ulysses} is known for the drastic
and deliberate stylistic shifts between the 18 chapters or ``episodes'' of which it is composed.
We used the text of the novel as an input, treating each episode as if it had been
composed by a different author.
We wanted to see whether these stylistic divisions could be automatically detected by a program.

Our input was Project Gutenberg's copy of a pre-1923 edition of the novel, released in July 2003.
After editing to remove some episode markers, the text has 32309 words.
The shortest episode is number 2, with 4414 words, and the longest is number 15, with 38161.


\begin{figure}[H]
\makebox[\textwidth][c]{\includegraphics[height=2.6in]{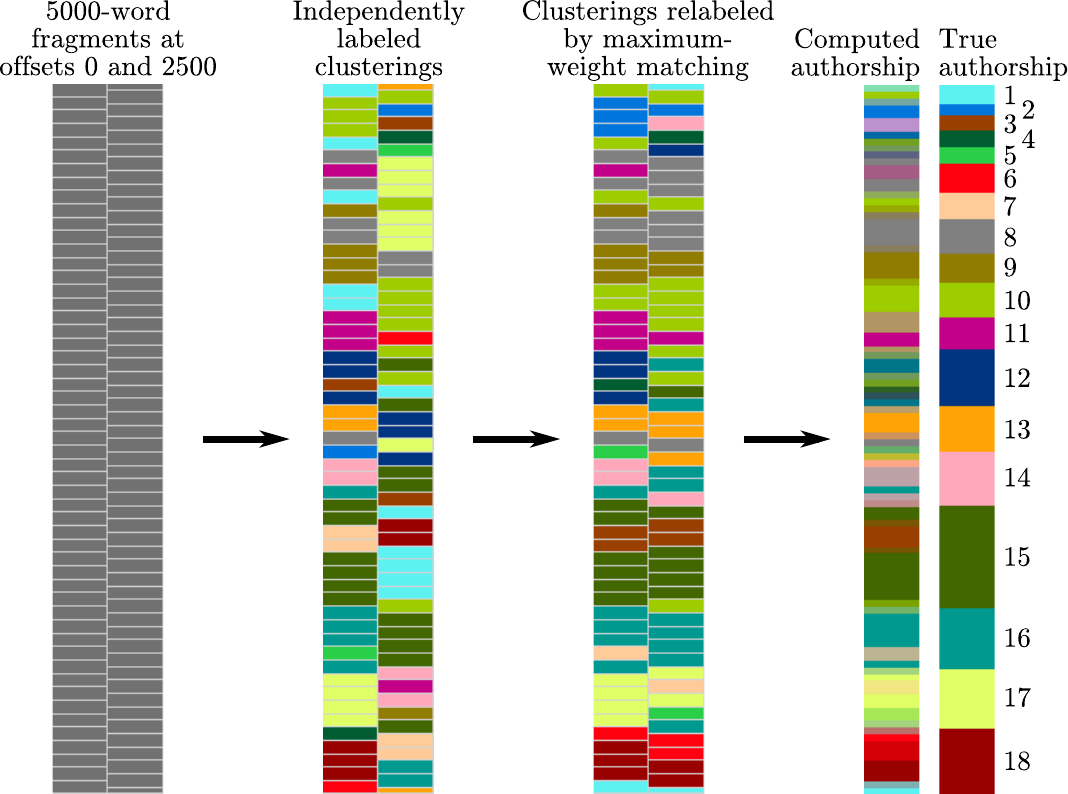}}
\caption{Attribution of \textsl{Ulysses}.
The agreement between the computed and true labels is $0.55$.}
\label{fig:ulysses}
\end{figure}

Figure~\ref{fig:ulysses} shows the process of attribution.
We followed the ``two clusterings, $n>2$ authors'' plan from Section~\ref{sec:two-n}.
We broke the input into uniform fragments of 5000 words, with one set of fragments
offset by 2500 words relative to the other.
The output is therefore a distribution over author labels for every 2500-word segment,
where the weights on each segment are either fully on one author or evenly split between two authors.
The two fragment sets are represented by the two gray columns at the left of the figure.
We then generated feature vectors for each 5000-word fragment using JStylo,
and clustered both fragment sets independently; this is the second pair of columns in the figure.
Author labels are shown as different colors.
Notice how there is hardly any agreement between the two columns.
We then computed an maximum-weight matching between the two fragment sets.
As shown in the third pair of columns, this makes many more labels agree.
On the right of the figure is the mean of the two relabeled clusterings,
aside the known true episode division.

We see that some episodes, perhaps those most stylistically distinct, are very closely matched:
8, 9, 10, 15, 16, and 17.
In other places the algorithm appears confused; reasons why may include stylistic similarity between episodes,
short episodes, or episode boundaries that do not fall close to a multiple of 2500 words.
The agreement between the computed authorship and the actual episode division is $0.55$
(random assignment would have $1/18\approx0.06$).

\subsection{The Homeric Hymn to Apollo}
\label{sec:apollo}

\begin{quote}
{
\obeylines
\greektext
μνήσομαι οὐδὲ λάθωμαι Ἀπόλλωνος ἑκάτοιο,
ὅντε θεοὶ κατὰ δῶμα Διὸς τρομέουσιν ἰόντα:
καί ῥά τ᾽ ἀναΐσσουσιν ἐπὶ σχεδὸν ἐρχομένοιο
πάντες ἀφ᾽ ἑδράων, ὅτε φαίδιμα τόξα τιταίνει.
}
\attrib{Lines 1--4 of the hymn}
\end{quote}

\noindent
The Homeric Hymns are a collection of ancient Greek hymns honoring the gods.
One of these, the Hymn to Apollo, believed to have been written before the 5th century B.C.,
is usually presented as one poem in two parts: a shorter ``Delian'' part and a longer ``Pythian'' part.
Textual and historical evidence suggests that, while traditionally combined,
they were originally composed independently.
Allen and Sikes~\cite{hymns} write:

\begin{quote}
The hymn to Apollo, in its present form, may be read as a continuous poem.
But the continuity lies only on the surface, and even the most casual reader cannot fail to be struck by
the abrupt transition at v.~179\ldots\ 
The unity of the hymn has been denied on artistic and literary
grounds.\ \ldots\ It therefore follows that the hymn is a compilation of
\emph{at least} two originally independent poems.
\end{quote}




Our goal was to see if a computer can automatically detect stylistic change between the two parts,
without a prior samples of writing from presumed authors.
We used as input the UTF-8 Greek text of the hymn~\cite{apollo}
from the Perseus Digital Library.
Including bracketed lines, the Greek hymn consists of 3840 words in 552 lines,
with 1241 words in the Delian portion and 2599 in the Pythian.

\begin{figure}[H]
\makebox[\textwidth][c]{\includegraphics[height=2.3in]{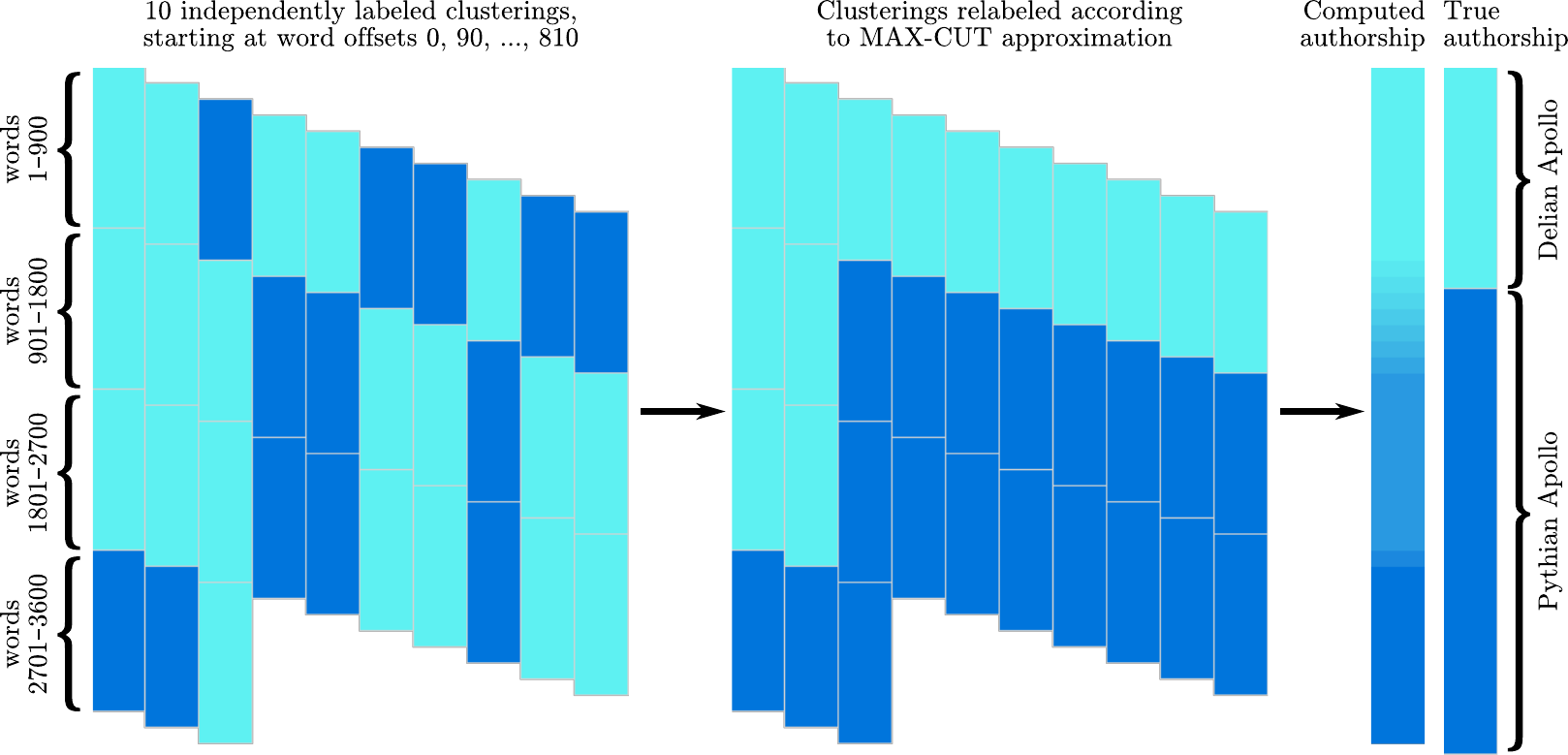}}
\caption{Attribution of the Hymn to Apollo.
The agreement between the computed and true labels is $0.84$.}
\label{fig:apollo}
\end{figure}

Figure~\ref{fig:apollo} illustrates the process of computing authorship.
We begin by breaking the input into fragments.
The input is rather short as authorship attribution goes---only 3736 words.
We use fragments of 900 words, and omit short fragments at the beginning and end,
meaning that each clustering will cluster only three or four fragments.
They start at offsets of 90 words relative to each other, so there are 10 clusterings.
The first clustering starts at offset $0$, over the four fragments consisting of words
1--900, 901--1800, 1801--2700, and 2701--3600.
The next clustering is over words 91--990, 991--1890, 1891--2790, 2791--3690, and so on.
In the figure, each column is an independently computed clustering.

The clustering may come out as in the left side of Figure~\ref{fig:apollo}.
Each fragment is colored according to its label assignment.
Because the algorithm is unsupervised, a light blue label stands for Unknown Author~1
and a dark blue label for Unknown Author~2. Because the clusterings are independent,
there is not much agreement between them.
The middle part of the figure shows the clusterings after relabeling by the MAX-CUT approximation.
For each clustering, the MAX-CUT approximation has freedom only to invert the labels, or not.
Here the algorithm has chosen to invert the labels of the third, sixth, seventh, ninth, and tenth clusterings for better agreement.
(Inverting the first, second, fourth, fifth, and eighth instead would result in the same cut and be just as good.)

The right side of Figure~\ref{fig:apollo} shows the mean of the relabeled clustering.
The balance of authorship may change as often as every 90 words,
making possible the smooth gradient-like transition seen here.
Comparing the computed authorship with the known division into Delian and Pythian parts,
we find them mostly in agreement, starting with one author at the beginning
and transitioning to another about 30\% through,
though it remains only 80\% sure of the second author until near the end of the text.
The agreement between the computed authorship and the actual Delian/Pythian split is $0.84$
(random assignment would have $0.5$).

\begin{figure}[H]
\makebox[\textwidth][c]{\includegraphics[height=2.3in]{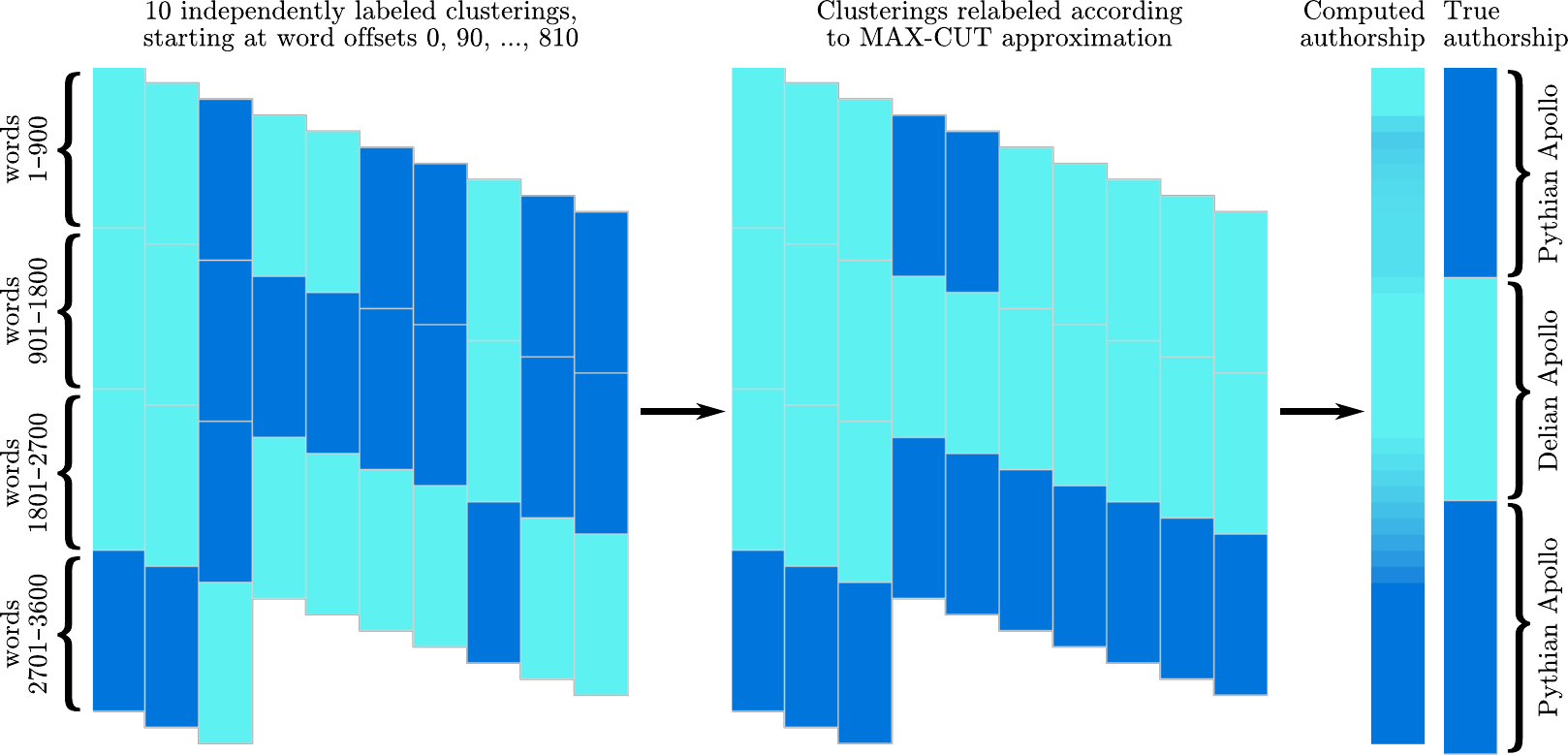}}
\caption{Attribution of a mangled Hymn to Apollo.
The Delian portion has been arbitrarily pasted into the middle of the Pythian.
The agreement is $0.72$.}
\label{fig:apollo-mixup}
\end{figure}

The mean assignment cannot have a sharp break between authors as the true authorship does;
it always takes at least 900 words to transition from one to another, during which transition it is on average at least half wrong.
Correspondingly an upper bound on the agreement achievable using this technique, in this case, is $(3726-450)/3726\approx0.88$.

We did another, artificial, test on the hymn.
In order to produce a true assignment that is non-contiguous,
we cut the Delian part and pasted it into the middle of the Pythian part.
The process is shown in Figure~\ref{fig:apollo-mixup}.
The second authorship change is very sharp but the first is less clear.
The total agreement with the known labeling is $0.72$.
Because there are two places where authorship changes, separated by
more than 900 words, the maximum agreement attainable by the algorithm is
$(3726-900)/3726\approx0.76$.

\subsection{Dimensionality reduction}
\label{sec:apollo-en}

In order to experiment with dimensionality reduction on feature vectors,
we need numeric feature vectors such as those from Section~\ref{sec:jstylo}.
We used JStylo to produce feature vectors from an English translation of the Hymn to Apollo~\cite{apollo-en}.
This text has 1768 words in the Delian portion and 3726 in the Pythian.


\begin{figure}[H]
\makebox[\textwidth][c]{\includegraphics[height=2.3in]{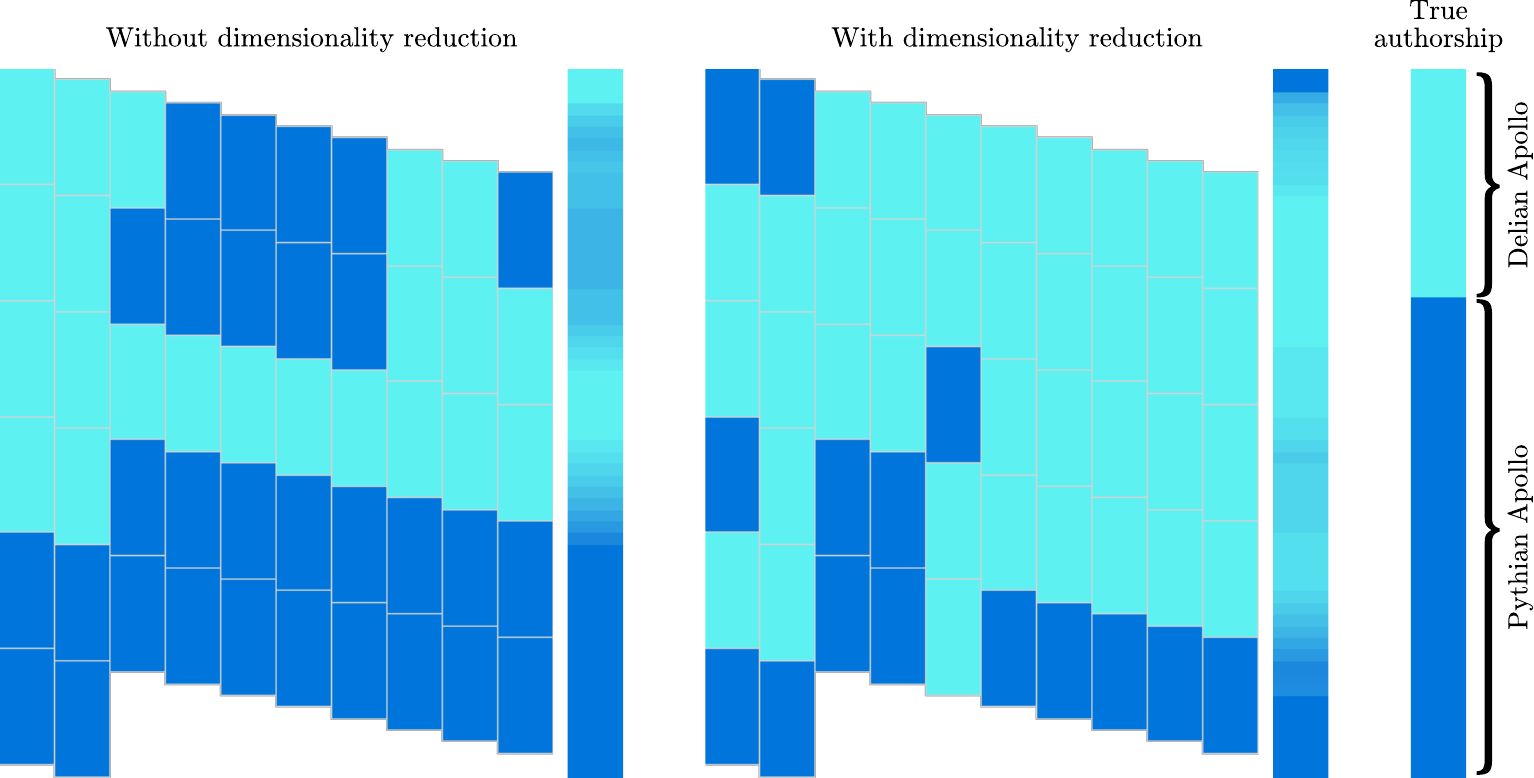}}
\caption{Before and after dimensionality reduction of an English translation of the Hymn to Apollo by random projection.
The agreement between the output labels is 0.56.}
\label{fig:dimreduc}
\end{figure}

Figure~\ref{fig:dimreduc} illustrates the difference with and without dimensionality reduction.
Notice that they agree mostly in the start and at the end, but they vary a bit more in the middle. This is presumably because the dimensionality reduction is an $2+\epsilon$ approximation.

\subsection{Source code}

Our source code and data files are available in a Git repository at
\url{https://repo.eecs.berkeley.edu/git-anon/users/fifield/cs270.git}.

\section{Future work}

We showed that the problem of matching multiple two-author clusterings reduces to MAX-CUT,
but didn't show that the problem is necessarily as hard as MAX-CUT.
The edge weights in the reduction have constraints; for example, they obey the triangle inequality.
We would like to determine whether the problem can be solved as well or better without reducing to MAX-CUT.

The case of $m>2$ clusterings and $n>2$ authors needs investigation to see how well it can be approximated.

\section{Acknowledgement}

We thank James Cook and Stephen Sansom for their comments on drafts of this report.

\bibliography{unsupauth}
\bibliographystyle{plain}

\end{document}